\documentclass{article}
\usepackage{graphicx} 
\usepackage[margin=0.68in]{geometry}
\usepackage{times}
\usepackage{microtype}
\usepackage{booktabs}
\usepackage{tabularx}
\usepackage{cite}
\usepackage[hidelinks]{hyperref}
\usepackage{enumitem}
\usepackage{etoolbox}
\setlist{nosep,leftmargin=*}
\AtBeginEnvironment{thebibliography}{\footnotesize\setlength{\itemsep}{0pt}}

\title{Machine Learning under Imperfect Data: Challenges and Methods}
\author{Masoumeh Zareapoor\\
Shanghai Jiao Tong University, China\\
\texttt{}}
\date{}

\begin{document}
\maketitle

\begin{abstract}
\noindent Machine-learning models are commonly developed under an assumption that training and test data are sufficiently complete, balanced, labelled, and drawn from compatible distributions. In practice, one or more of these conditions is often violated. Measurements may be missing or corrupted, rare classes may be poorly represented, supervision may be weak, and the deployment environment may differ from the training environment. These imperfections are usually treated as separate technical problems, although they alter learning through a small number of shared mechanisms: loss of information, biased empirical risk, ambiguous supervision, and unstable representations. This short survey organises representative methods around these mechanisms. It reviews reconstruction and generation, rebalancing and representation calibration, learning with limited supervision, adaptation across domains and modalities, and reliability under distribution change. The discussion highlights the limits of plausible reconstruction, benchmark-specific correction, and adaptation without trustworthy feedback. It concludes with directions for evidence-aware learning, uncertainty-preserving prediction, and evaluation that separates visual plausibility from decision utility.
\end{abstract}

\section{Introduction}
The success of modern machine learning has been supported by large datasets, scalable optimisation, and expressive architectures such as residual networks and transformers~\cite{he2016deep,vaswani2017attention,dosovitskiy2021image}. Yet data collected outside controlled benchmarks rarely satisfy the assumptions under which these models are trained. Sensors fail, observations are occluded, labels are expensive, class frequencies are uneven, and future inputs depart from the training distribution. We use \emph{imperfect data} as a term for observations whose information content, supervision, sampling distribution, or deployment correspondence is insufficient for the intended prediction task.

This definition is deliberately task dependent. A low-resolution image may be adequate for scene recognition but inadequate for small-object localisation. Likewise, a dataset can be large while providing little evidence about rare classes or unfamiliar environments. Imperfection should therefore not be identified only with low data quality, it is a mismatch between the available evidence and the decision that a model is expected to make.

The literature addresses this mismatch through several connected strategies. Reconstruction methods infer absent content from observed context~\cite{pathak2016context,shamsolmoali2024distance}, generative models learn a distribution over plausible observations~\cite{goodfellow2014generative,ho2020denoising, shamsolmoali2022gen, rombach2022high, zareapoor2018advance}, imbalance-aware methods change sampling, losses, or representations~\cite{lin2017focal}, weakly supervised methods extract structure from coarse annotations~\cite{zhou2016learning} and adaptation methods reduce discrepancies between training and deployment domains~\cite{ganin2016domain}. These approaches are reviewed here through the type of evidence they repair and the uncertainty they introduce.

\section{Forms of Data Imperfection}
Table~\ref{tab:taxonomy} presents a compact taxonomy. The categories overlap: an infrared image may be complete in its own modality but incomplete with respect to colour and texture, a weakly labelled sample may also belong to a rare class, and domain shift may change both appearance and class frequency. The distinction is nevertheless useful because each condition changes the learning problem in a different way.

\begin{table*}[t]
\centering
\small
\caption{Common forms of imperfect data and representative responses.}
\label{tab:taxonomy}
\begin{tabularx}{\textwidth}{@{}p{2.4cm}p{3.6cm}Xp{3.2cm}@{}}
\toprule
Imperfection & Learning consequence & Representative responses & Central evaluation question \\
\midrule
Missing or degraded content & Multiple latent observations may explain the same input & Super-resolution, inpainting, completion, conditional generation, multimodal fusion & Is the recovery faithful, or merely plausible? \\
Imbalanced sampling & Empirical risk is dominated by frequent groups or easy examples & Reweighting, resampling, focal losses, distribution calibration, synthetic augmentation & Does minority performance improve without distorting calibration? \\
Limited supervision & Labels do not uniquely specify the target structure & Multiple-instance learning, pseudo-labels, consistency, distillation, region-word alignment & Does the model discover the intended evidence or a shortcut? \\
Distribution shift & Training correlations are not stable at deployment & Invariant features, domain adaptation, source-free adaptation, domain generalisation & Are gains stable across genuinely unseen shifts? \\
Modality mismatch & Evidence has different statistics or is absent in one channel & Shared embeddings, privileged information, contrastive alignment, fusion & Is complementary information preserved without forced alignment? \\
\bottomrule
\end{tabularx}
\end{table*}

Missingness reduces information about the target, whereas imbalance changes how that information is represented in the objective. Limited supervision creates ambiguity between observed labels and latent explanatory structure. Distribution and modality shifts alter which learned relations continue to hold. A robust method must identify which of these mechanisms is present, applying a generic augmentation or larger model may conceal the symptom without resolving the mismatch.

\subsection{A Learning-Theoretic View}
Let $x$ denote the latent complete observation, $\tilde{x}$ the evidence available to the learner, $y$ the prediction target, and $e$ the environment in which the sample is acquired. Standard empirical risk minimisation implicitly treats the observed triples as adequate samples from the deployment distribution. Imperfection breaks this correspondence in several ways. A degradation process may map many values of $x$ to the same $\tilde{x}$, a sampling process may change the observed class prior, a labelling process may reveal only a coarse or noisy version of $y$, and a change from training environment $e_s$ to deployment environment $e_t$ may alter the relation between observations and targets.

This view separates three questions that are sometimes conflated. \emph{Identifiability} asks whether the desired target can, even in principle, be recovered from the available evidence. \emph{Learnability} asks whether the data and hypothesis class are sufficient to estimate the relevant relation. \emph{Robustness} asks whether that relation continues to support decisions when the acquisition or deployment conditions change. More capacity can improve learnability while leaving identifiability unchanged. For example, no deterministic inpainting network can recover the unique original content of a fully occluded region when several completions agree with the visible context.

The distinction also clarifies the role of inductive bias. Convolutional structure favours spatial locality, transformers support long-range interaction, geometric representations preserve selected transformations, and pretrained models import statistical regularities from external data. These priors reduce the range of solutions compatible with incomplete evidence, but they do not create new observations. Their value depends on whether the imported assumptions match the target environment. A useful method should therefore state which information is observed, which is inferred from training regularities, and which is supplied through architecture or pretraining.

\section{Learning from Incomplete Evidence}
Early deep reconstruction systems learned direct mappings from degraded inputs to high-quality outputs. Dense and progressive feature reuse improved single-image super-resolution, while adversarial training encouraged perceptually realistic detail~\cite{goodfellow2014generative}. Similar ideas extend to depth enhancement and multispectral fusion, where information is transferred across scales or sensing channels. These methods demonstrate that the form of degradation matters: missing regions, and absent modalities require different conditioning signals.

Image completion makes the ambiguity explicit because an unobserved region can admit many valid solutions. Context encoders established adversarial feature learning for filling holes~\cite{pathak2016context}, transformer-based approaches later modelled longer-range dependencies and adapted contextual information to the mask geometry~\cite{shamsolmoali2024distance}. Diffusion models provide iterative conditional generation and strong sample quality~\cite{ho2020denoising,rombach2022high}. Recent completion work uses time-aware conditioning to reduce unnecessary diffusion steps and context-adaptive generation to coordinate structure and detail~\cite{mayet2025td,shamsolmoali2025missing}. For 3-D data, sparse or partial observations introduce geometric constraints, autoregressive and discrete-diffusion formulations model shape structure beyond a regular image grid.

The main limitation is that perceptual quality does not establish evidential correctness. A generated region can be visually convincing but unsupported by the input. Pixel metrics penalise valid alternatives, while perceptual metrics may reward confident fabrication. Evaluation should therefore combine fidelity on observed regions, conditional diversity in unobserved regions, and performance on an independently defined downstream task. In consequential applications, the model should represent multiple possible completions or abstain when the evidence is insufficient.

\section{Sampling and Limited Supervision}
Class imbalance shifts optimisation toward frequent classes. Loss reweighting and focal loss reduce the influence of easy majority examples~\cite{lin2017focal}, whereas data-level methods increase exposure to minority samples. Adversarial oversampling can generate minority observations, but synthetic examples may reproduce biases or occupy regions unsupported by real data. Estimating minority statistics and calibrating their representation offers an alternative to unconstrained generation~\cite{shamsolmoali2023vtae, yao2021depth}. The choice between these strategies depends on whether the deficit is sample quantity, coverage of intra-class variation, or separation at the decision boundary.

Limited supervision concerns the granularity or reliability of labels. Multiple-instance learning infers instance-level evidence from bag-level annotations, adversarial formulations can encourage diverse pose hypotheses when precise labels are scarce~\cite{shamsolmoali2020amil}. Attention distillation similarly transfers localisation information without dense annotation. In open-vocabulary detection, region--word alignment replaces a fixed class set with semantic supervision, but imperfect correspondence between visual and linguistic spaces can introduce a new form of label ambiguity.

Pseudo-label and consistency methods exploit unlabelled data effectively~\cite{sohn2020fixmatch}, although confirmation bias can turn early errors into supervision. A common weakness across weak supervision and imbalance correction is that aggregate accuracy hides where the evidence fails. Class-wise recall, calibration, localisation quality, and subgroup performance should be reported together. Ablations should also separate improvement due to additional samples from improvement due to the learning mechanism.

\section{Distribution and Modality Shifts}
Domain adaptation assumes that labelled source data and target data share a task but differ in distribution. Domain-adversarial learning seeks features that support the task while obscuring domain identity~\cite{ganin2016domain}. In crowd counting, synthetic supervision and edge-aware adaptation can transfer structural cues to real scenes. Domain generalisation instead aims to prepare a model for unseen domains without target data, multi-step bidirectional learning has been used to address appearance variation in visible-infrared person re-identification.

Cross-modal learning is not simply another domain-alignment problem. Modalities may be complementary rather than interchangeable. Privileged intermediate information can guide a deployable branch during training, and bidirectional contrastive objectives can align paired modalities while retaining direction-specific structure~\cite{zareapoor2025bimac}. Excessive invariance may discard precisely the information that makes a modality useful. Alignment should therefore be selective: shared semantics can be brought together, while modality-specific uncertainty and detail are preserved.

The same caution applies to geometric invariance and equivariance. Remote-sensing objects appear across scale, orientation, and viewpoint, so rotation-equivariant pyramids and feature distillation can improve detection from sparse or small visual evidence~\cite{shamsolmoali2023efficient}. Yet invariance is appropriate only when a transformation does not change the label. Orientation may be irrelevant for identifying a vehicle but essential for estimating its direction of travel. Imperfect-data learning therefore benefits from transformations encoded according to the task, rather than from indiscriminate removal of variation.

Distribution shift also arises over time. Continual learning must acquire a new task without overwriting parameters that support earlier tasks. Proximal decoupling separates task switching from destructive updates, illustrating a broader principle: adaptation should constrain changes according to the evidence available at the time of update. Benchmarks such as WILDS emphasise evaluation across naturally occurring shifts~\cite{koh2021wilds}, nevertheless, averaged benchmark results cannot guarantee robustness to shifts outside the chosen test environments.

\section{Reliability under Imperfect Evidence}
Reliability requires more than improving average predictive accuracy. Adversarial perturbations expose sensitivity to small, targeted changes, fast adversarial training is computationally attractive but can suffer catastrophic overfitting, motivating optimisation strategies that separate conflicting update effects~\cite{zareapoor2024rethinking}. Out-of-distribution detection and predictive uncertainty attempt to identify inputs for which learned confidence is unreliable~\cite{hendrycks2017baseline,ovadia2019can}. 

Interpretability can reveal whether a model relies on suitable evidence. Class activation mapping localises influential regions~\cite{zhou2016learning}, while part-prototype models associate predictions with human-inspectable components. Such explanations must be evaluated for faithfulness as well as visual appeal. Efficiency is also relevant: sparse mixture-of-experts models route inputs through a subset of parameters, but routing errors can amplify data imbalance and specialise experts around spurious partitions. Reliability therefore concerns the entire learning data selection, representation, optimisation, routing, and confidence, rather than a final uncertainty score.

\section{Evaluation Principles}
Evaluation under imperfect data should establish what information a method uses and how its benefits change as evidence deteriorates. A single test condition is insufficient because it cannot distinguish a useful mechanism from tuning to one corruption level, class ratio, annotation regime, or domain pair.

\noindent \textbf{Controlled severity.} Missing-region size, spatial resolution, label frequency, imbalance ratio, and domain discrepancy should be varied systematically. Performance curves across severity levels are more informative than a single average. They reveal thresholds at which a method becomes unstable and whether its advantage grows in the condition it is designed to address.

\textbf{Matched comparisons.} Data-level methods should be compared at equal amounts of real supervision and, where possible, equal computational budgets. Synthetic augmentation must be separated from the effect of additional training iterations. Adaptation methods should report results before and after adaptation and include a no-adaptation reference, since a strong pretrained representation may account for much of the observed robustness.

\textbf{Multiple dimensions of performance.} Overall accuracy is inadequate when classes or groups are imbalanced. Macro-averaged measures, class-wise sensitivity, calibration, and worst-group performance expose different failure modes. For reconstruction, distortion and perceptual measures should be complemented by structural consistency and downstream task performance. For open-set or shifted inputs, selective risk and coverage show whether abstention meaningfully reduces error.

\textbf{Uncertainty and validity.} When the evidence admits several targets, evaluation should not assume that one reference is the only acceptable output. Conditional generation can be assessed through fidelity to observed content, diversity among completions, and consistency with known constraints. Conversely, diversity should not excuse outputs that contradict the evidence. Human evaluation can be useful for perceptual quality, but it should not substitute for task-specific validity checks.

\textbf{Leakage-resistant protocols.} Hyperparameters, stopping rules, and model selection must not use information from the nominally unseen test environment. This is particularly important in domain generalisation and source-free adaptation, where repeated evaluation on the target can quietly turn an unseen-domain experiment into supervised tuning. Reporting variability across seeds and multiple domains is also necessary because adaptation can produce unstable gains.

These principles encourage claims tied to observable evidence. A model should not be described as robust merely because it improves one shifted benchmark, nor should a visually convincing completion be treated as recovery of the unknown ground truth. Evaluation is strongest when it directly tests the mechanism proposed by the method.

\section{Future Directions}
Several priorities follow from this unified view.

\noindent \textbf{Evidence-aware objectives.} Models should distinguish observed information from inferred content. Reconstruction and multimodal systems need objectives that reward conditional support, not only visual realism. Producing sets or distributions of predictions may be more honest than selecting one completion.

\noindent \textbf{Compositional imperfections.} Most benchmarks isolate one problem, yet missingness, imbalance, noisy supervision, and shift often co-occur. Evaluation should vary these factors jointly and report which combinations cause failure. This would also reveal whether a method corrects a mechanism or overfits a benchmark convention.

\noindent \textbf{Adaptation with safeguards.} Test-time and continual adaptation require criteria for deciding when new evidence is sufficient to justify an update. Conservative updates, reversible memory, uncertainty thresholds, and explicit monitoring of earlier capabilities could reduce silent degradation.

\noindent \textbf{Task-grounded evaluation.} Image quality, feature alignment, and aggregate accuracy are incomplete proxies. Evaluation should test downstream decisions, minority and subgroup outcomes, calibration under shift, and the faithfulness of explanations. Where the input does not determine a unique answer, benchmarks should recognise multiple valid outputs.

\noindent \textbf{Selective use of foundation models.} Large pretrained vision and language models provide broad priors, as illustrated by contrastive vision-language learning~\cite{radford2021learning} and transformer representations~\cite{dosovitskiy2021image}. These priors can compensate for limited local data, but they may also import undocumented biases. Adaptation should preserve useful general knowledge while exposing uncertainty about populations or modalities that were poorly represented during pretraining.

\section{Conclusion}
Imperfect data are not a peripheral exception to machine learning, they describe the conditions under which many systems operate. Missing content, biased sampling, limited labels, domain changes, and modality differences create distinct forms of insufficient evidence. Reconstruction, rebalancing, weak supervision, adaptation, and robust learning provide useful responses, but each can introduce new uncertainty. Progress therefore depends on matching the method to the mechanism of imperfection and evaluating whether predictions are supported by the available evidence. Models that communicate uncertainty, adapt conservatively, and are tested under interacting imperfections offer a more credible route toward dependable learning in open environments.

\bibliographystyle{unsrt}
\bibliography{main}

\end{document}